\documentclass[10pt,twocolumn,letterpaper]{article}

\usepackage[pagenumbers]{wacv} %

\usepackage{graphicx}
\usepackage{amsmath}
\usepackage{amssymb}
\usepackage{booktabs}
\usepackage{pifont}
\usepackage{multirow}
\usepackage{booktabs}
\usepackage{makecell}
\usepackage[scientific-notation=true]{siunitx}
\usepackage[dvipsnames]{xcolor}

\usepackage[pagebackref,breaklinks,colorlinks]{hyperref}

\usepackage[capitalize]{cleveref}
\crefname{section}{Sec.}{Secs.}
\Crefname{section}{Section}{Sections}
\Crefname{table}{Table}{Tables}
\crefname{table}{Tab.}{Tabs.}

\def\wacvPaperID{1981} %
\def\confName{WACV}
\def\confYear{2024}

\begin{document}

\title{CaSAR: Contact-aware Skeletal Action Recognition}

\author{Junan Lin$^1$*~~~~~~~~~
Zhichao Sun$^1$*~~~~~~~~~
Enjie Cao$^1$*~~~~~~~~~
Taein Kwon$^1$~~~~~~~~~\\
Mahdi Rad$^2$~~~~~~~~~
Marc Pollefeys$^{1,2}$
\smallskip 
\\
$^1$ETH Z\"urich~~~~~~$^2$Microsoft MR \& AI Lab, Z\"urich
}

\maketitle
\let\thefootnote\relax\footnotetext{*Co-first authors}

\begin{abstract}

  Skeletal Action recognition from an egocentric view is important for applications such as interfaces in AR/VR glasses and human-robot interaction, where the device has limited resources. Most of the existing skeletal action recognition approaches use 3D coordinates of hand joints and 8-corner rectangular bounding boxes of objects as inputs, but they do not capture how the hands and objects interact with each other within the spatial context. In this paper, we present a new framework called Contact-aware Skeletal Action Recognition  (CaSAR). It uses novel representations of hand-object interaction that encompass spatial information: 1) contact points where the hand joints meet the objects, 2) distant points where the hand joints are far away from the object and nearly not involved in the current action. Our framework is able to learn how the hands touch or stay away from the objects for each frame of the action sequence, and use this information to predict the action class. We demonstrate that our approach achieves the state-of-the-art accuracy of $91.3\%$ and $98.4\%$ on two public datasets, H2O and FPHA, respectively.

\end{abstract}

\section{Introduction}

Action recognition from an egocentric view has become increasingly important in recent years, particularly for real-world applications where hands interact with an object, such as interfaces in augmented reality and virtual reality and human-robot interaction,  \cite{8702353, 10.1007/978-3-319-95270-3_21, MANO:SIGGRAPHASIA:2017}. Light-weight methods and accurate recognition of the actions are critical for providing a seamless and intuitive user experience. However, existing light-weight egocentric action recognition, such as skeletal action recognition, approaches have limitations due to missing hand-object relation in the existing representation of the hands and objects \cite{OpenPose, MANO:SIGGRAPHASIA:2017, Mousavian_2017_CVPR}. These methods mostly use 3D coordinates of hand joints and 8-corner rectangular bounding box of objects as inputs \cite{Tekin_2019_CVPR, kwonH2OTwoHands2021}, which do not consider the interaction information between hands and the object. This lack of information can result in inaccurate results, and therefore, will not be practical for the real-world scenarios.

\begin{figure}
    \centering
    \includegraphics[width=0.48\textwidth]{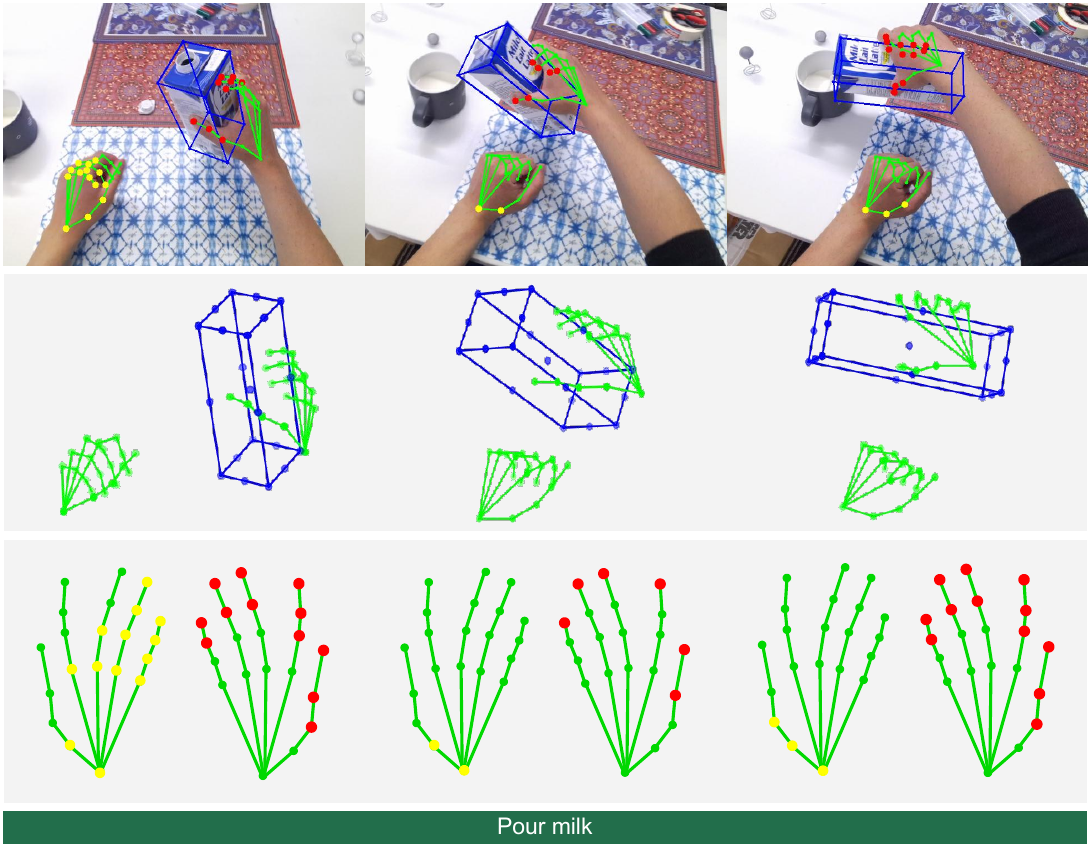}
    \caption{\textbf{Action Recognition Using 3D Hand and Object Pose Representations.} We propose Contact-aware Skeletal Action Recognition (CaSAR): a new framework for skeletal action recognition using a novel 3D hands and object poses representation, which includes hand joints positions (green), object 8-corner bounding box (blue), as well as contact points (red) and distant points (yellow) as interaction information. (Row 1) Image frame, (Row 2) Input skeleton frame, (Row 3) Contact points and distant points, (Row 4) Estimated action classes.} %
    
    \label{fig:intro}
\end{figure}

Very recently, \cite{Cho_2023_CVPR} illustrated that having the explicit relational information between hands and objects provides a strong cue for the recognition task. To obtain this explicit information, they propose to compute contact maps between hands and object meshes, given 3D hands and object poses. This, however, heavily relies on having accurate and complete mesh models of the hands and the objects, which can be impractical for all possible objects of interest. To address this issue, we propose a novel skeletal action recognition approach called Contact-aware Skeletal Action Recognition (CaSAR). The key idea is to enforce the network to learn the hands-object interaction information, contact-map, given only hand joint positions and object 8-corner bounding box for each frame of the action sequence. Then, the model learns to predict the action class of the corresponding sequence based on hands and object poses as well as the predicted contact-map, as shown in Figure \ref{fig:intro}. Our contact-map representation not only encompasses the contact points where hand joints are in direct contact with the object, but also includes distant points for joints located beyond a specified threshold to the object, as illustrated in Figure \ref{fig:intro}. By incorporating this additional distant information, our method demonstrates even further improvements in accuracy, as the network gains a better understanding of spatial relations between the hands and the object, resulting in enhanced performance on the given task. Moreover, our proposed contact-aware approach does not require additional ground-truth contact information, nor hand and object meshes at inference time.
To evaluate the performance of our proposed method, we conducted experiments on two public datasets of H2O \cite{kwonH2OTwoHands2021} and FPHA \cite{Garcia-Hernando_2018_CVPR}. Our experiments show that our proposed contact-aware approach achieves state-of-the-art accuracy of $91.3\%$ and $98.4\%$, on H2O and FPHA, respectively. These results demonstrate the effectiveness of our proposed approach in improving the accuracy of egocentric action recognition.

The contributions of this paper are as follows. 
First, we propose a novel contact-aware skeletal action recognition framework, CaSAR, that can learn hands-object spatial relations explicitly from hand joints and object bounding boxes as well as geometric shapes implicitly, which removes the need of 3D meshes during the testing time. As a result, our method can adapt to in-the-wild applications.
Second, we introduce a novel concept, contact-map, which can model hand-object spatial relations explicitly by setting contact points where the hand joints meet the objects and distant points where the hand joints are far away from the object. Moreover, learning contact-map leads to significant improvement in the accuracy for the task of action recognition.
Third, we conduct experiments on two public datasets to evaluate the performance of our proposed method, and achieve state-of-the-art accuracy on both datasets.
Overall, our proposed method has significant potential for improving the accuracy of egocentric action recognition and also making the action recognition model understand contact and distant information, which could have important implications for a wide range of real-world applications, including human-computer interaction, augmented reality, and virtual reality.

\section{Related Work}
In this section, we focus on the following research areas closely related to our work: 1) hand-object interaction recognition, 2) skeletal action recognition, and 3) hand-object contact relation.

\begin{figure*}
    \centering
    \includegraphics[width=\textwidth]{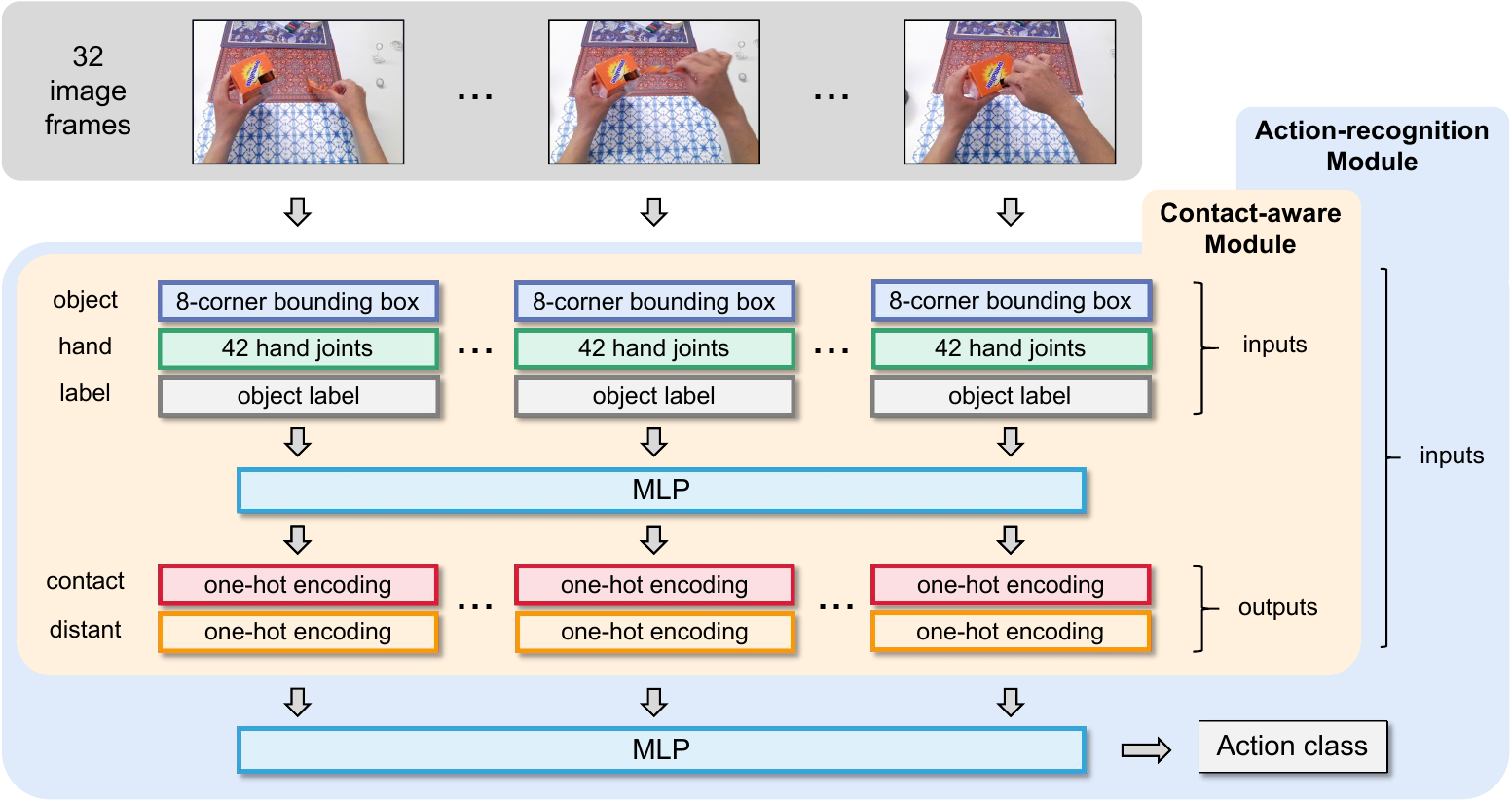}
    \caption{The overall architecture of CaSAR model. We choose 32 as the number of extracted frames for each video segment. For each frame, in addition to the 8-corner bounding box as object pose representation and 42 hand joints as hand pose representation, we further integrate a 42-element one-hot encoding vector that encapsulates contact information and another 42-element one-hot encoding vector indicating distant information, which allows the model to incorporate critical spatial interaction. }
    
    \label{pipeline}
\end{figure*}

\subsection{Hand-object Interaction Recognition}
General action recognition is mainly performed by convolution-based and transformer-based neural networks, which extract RGB-based features. Some examples include I3D \cite{carreira2018quo}, SlowFast \cite{feichtenhoferSlowFastNetworksVideo2019a} and Two-stream \cite{Shi_2019_CVPR}. In terms of hand-object interaction, earlier works on the hand-object interaction recognition task \cite{Bambach_2015_ICCV, fathi, fathi2, Ma_2016_CVPR} extracted similar appearance features to recognize the hand-object interaction in the egocentric viewpoint. Recently, Garcia-Hernando et al. \cite{Garcia-Hernando_2018_CVPR} showed that hand poses are more helpful than RGB images in recognizing hand actions. Tekin et al. \cite{Tekin_2019_CVPR}, Kwon et al. \cite{kwonH2OTwoHands2021}, and Cho et al. \cite{Cho_2023_CVPR} proposed each unified framework that performs pose estimation of hands and objects and the hand-object interaction recognition via Long Short-Term Memory (LSTM) \cite{Hochreiter}, Graph Convolutional Network (GCN) \cite{kipf2016semi}, and Transformer \cite{vaswani2017attention}, respectively. In our paper, we present a simple but effective Multilayer Perceptron (MLP) \cite{haykin1994neural} framework with contact context for the same task.

\subsection{Skeletal Action Recognition}

 Skeletal action recognition models have been extensively developed nowadays thanks to recent datasets such as NTU120~\cite{liu2019ntu} and H2O~\cite{kwonH2OTwoHands2021} that include human skeletons. Some popular model structures are recurrent neural network (RNN) \cite{rumelhart1985learning} and GCN. Du et al. proposed hierarchical RNN \cite{Du_2015_CVPR}, which divides human skeleton into five parts and trains five sub-RNNs for action recognition. Li et al. introduced AS-GCN \cite{liActionalStructuralGraphConvolutional2019}, which extracts implicit relations among joints to improve the performance of traditional GCN which only makes use of fixed skeleton graphs. Zhang et al. came up with Semantics-Guided Neural Networks (SGN) \cite{Zhang_2020_CVPR}, which combines GCN and convolution neural network (CNN) \cite{fukushima1980neocognitron} and introduces high-level semantics of joints to enhance the feature representation capability. While skeleton-based action recognition has been extensively researched, little focus has been paid to skeletal hand-object interaction recognition, in which the object and the interaction between hands and the object play a significant role. In our paper, hand-object interaction is tailored for this skeletal action recognition.

\subsection{Hand-object Contact Relation}
In the realm of action recognition, hand-object contact relations emerged as a powerful tool for encoding dynamic interactions. Brahmbhatt et al. \cite{brahmbhattContactDBAnalyzingPredicting2019a} proposed to use thermal cameras to capture the hand-object contact maps, reflecting the commonly contacted regions of the object following the grasping action. They additionally presented ContactPose \cite{brahmbhattContactPoseDatasetGrasps2020a}, one of the first datasets that combine hand-object contact information with hand pose, object pose, and RGB-D images. Very recently, Cho et al. \cite{Cho_2023_CVPR} also combined contact relations which encode the interaction between two hands
and the object into their Transformer-based Unified Recognition module. 
Unlike the previous work, our framework considers both contact and distant information, which makes the network learn 3D geometric information and leads to better action recognition performance.

\section{Method}

In our work, we first define a novel concept of contact-map, which includes contact points and distant points. We then introduce our framework as illustrated in Figure~\ref{pipeline}, CaSAR, which leverages the 3D spatial information that is explicitly provided for the task of skeletal action recognition. In this work, we provide contact-map as 3D context. We describe each part in this section.

\subsection{Contact-map}
\begin{figure*}
    \centering
    \includegraphics[width=\textwidth]{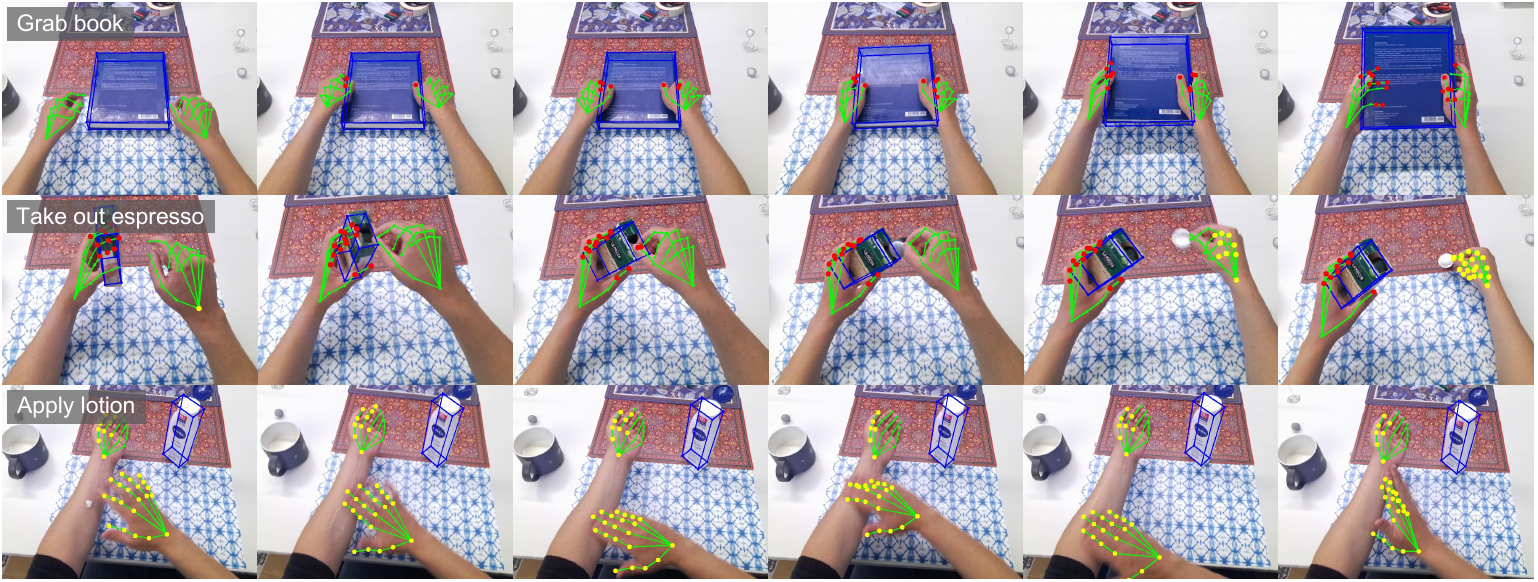}
    \caption{Examples of Contact points (red) that are closer than 2~cm to the object and distant points (yellow) that are more than 20~cm away from the object on the H2O dataset.}
    \label{contact_distant}
\end{figure*}

Contact-map encompasses spatial information as it consists of contact points, where the hand joints meet the objects, together with distant points, where the hand joints are far away from the objects. The literature has discussed the importance of contact points in hand-object interaction \cite{Cho_2023_CVPR} to some extent, but there has been little attention given to the role of distant points. As we show in our experiment section and the ablation study, our novel formulation of contact-map for hand-object interaction that also focuses on the significance of distant points performs better compared to contact points only. Figure \ref{contact_distant} illustrates contact and distant points.

\medskip
\noindent{\textbf{Contact Points.}} \qquad We categorize each hand joint as a contact point if it is in contact with the object mesh. In practice, a joint is considered as a contact point if its Euclidean distance to the closest object mesh vertex is less than a threshold $\eta_c$. In our experiments, we use $\eta_c = 2~cm$, which is an optimal value considering the object dimension and sparsity of the object mesh.

\medskip
\noindent{\textbf{Distant Points.}} \qquad  A hand joint is categorized as a distant point if its Euclidean distance to the closest object mesh vertex surpasses a threshold of $\eta_d$.  In our experiments, we use $\eta_d = 20~cm$ for the H2O dataset and  $\eta_d = 10~cm$ for the FPHA dataset considering their hand movement distribution,
which helps to obscure the joints that have little interaction to the object.

\subsection{CaSAR}
\label{sec:CaSAR}
Skeletal action recognition methods aim to understand and classify human hand-object interaction actions from 3D hand and 3D object pose. In this work, we propose to leverage 3D spatial context in addition to the skeleton data. We use contact and distant points as the context.

Let $\mathcal{T} = \{(x_1, y_1), (x_2, y_2), \ldots, (x_n, y_n)\}$ be the skeleton training set. In this set, each $x_i$ is composed of $N_f$ frames of the following parts: a sequence of 3D coordinates of 21 hand joints on both hands denoted as $\mathbf{h}_j \subset \mathbb{R}^{3 \times 21 \times 2}$, a sequence of 3D object poses represented by 3D coordinates of 21 points on an 8-corner bounding box denoted as $\mathbf{o}_j \subset \mathbb{R}^{3 \times 21}$, and a sequence of object label represented using an 8-element one-hot encoding vector denoted as $\mathbf{l}_j \subset \mathbb{R}^{8}$, in which $j=1,2,\ldots,{N_f}$. The corresponding action label is denoted as $y_i$. Each $x_i \subset \mathbb{R}^{{N_f} \times 197}$ is represented as:
\[
x_i = \underbrace{\mathbf{h}_1\mathbf{o}_1\mathbf{l}_1}_{\text{hand, object, object label}}
\ldots\underbrace{\mathbf{h}_{N_f}\mathbf{o}_{N_f}\mathbf{l}_{N_f}}_{\text{hand, object, object label}}
\]
where $N_f = 32$ is the total number of frames per action. Notice that 
$\mathbf{l}_1=\mathbf{l}_2=\ldots=\mathbf{l}_{N_f}$ , as the object label remains the same within the same action.

We train the Action Recognition network denoted as $g$ on the skeleton training set $\mathcal{T}$, where the input is concatenated with the contact-map predicted by network $f$. Network $f$ is trained on a contact training set $\mathcal{T^C} = \{(p_1, q_1), (p_2, q_2), \ldots, (p_m, q_m)\}$, where ${p_i} = [\mathbf{h}_i, \mathbf{o}_i, \mathbf{l}_i] \subset \mathbb{R}^{197}$ consists of 3D hand joints, 3D object pose, and object label, and ${q_i} = [\mathbf{c}_i, \mathbf{d}_i] \subset \mathbb{N}^{84}$ represents contact points $\mathbf{c} \subset \mathbb{N}^{42}$ and distant points $\mathbf{d} \subset \mathbb{N}^{42}$ of both left and right hands.  $i$ denotes the i-th sample in the dataset.

\noindent{\textbf{Loss.}}
We optimize the following objective function over $\theta_f$ and $\theta_g$, the parameters of networks $f$ and $g$ as follows: 
\begin{equation}
\label{eq:loss_total}
\mathcal{L}(\theta_g, \theta_f;\mathcal{T}, \mathcal{T}^C) = \mathcal{L}_{\text{contact}} + \lambda \mathcal{L}_{\text{action}} \qquad
\end{equation}
Where $\lambda$ is a hyperparameter that controls the balance between the action recognition and contact map prediction objectives.  $\mathcal{L}_{\text{action}}$ is cross-entropy action recognition loss associated with network $g$: 
\begin{equation}
\begin{aligned}
\mathcal{L}_{\text{action}} = -\frac{1}{n} \sum_{(x_i, y_i) \in \mathcal{T}} \sum_{c=1}^{C}  y_{i,c}  \cdot \log(g([x_i, f(x_i;\theta_f)]_c;\theta_g))
\end{aligned}
\end{equation}
where $y_{i,c}$ is the indicator function that equals 1 if sample $i$ belongs to class $c$, otherwise 0. $g(\cdot)_c$ is the predicted probability that sample 
$i$ belongs to class $c$, and $C$ is total number of action classes.

$\mathcal{L}_{\text{contact}}$ is the contact-map prediction loss associated with network $f$. As in practice, the contact-map contains imbalanced class distribution between contact points and non-contact points, and distant points and non-distant points; we use Focal Loss \cite{lin2017focal} for $\mathcal{L}_{\text{contact}}$:

\begin{equation}
\begin{aligned}
\mathcal{L}_{\text{contact}} = -\frac{1}{m} \sum_{(p_i, q_i) \in \mathcal{T}^C}  \alpha q_i \cdot (1 - \hat{q}_i)^{\gamma} \cdot \log(\hat{q}_i) + \\(1 - \alpha) (1- q_i) \cdot (\hat{q}_i)^{\gamma} \cdot \log(1 - \hat{q}_i) 
\end{aligned}
\end{equation}
where $\hat{q}_i = h(p_i;\theta_h)$ is the predicted probability of the positive class for sample $i$.
In our experiments, we use $\alpha=0.5$ and $\gamma=4$.

\section{Experiments}
In this section, we present and discuss the results of our evaluation on two public datasets. We first describe the datasets and the implementation details. Then, we evaluate our method and compare it to the state-of-the-art as well as the results of an ablative analysis of our method.

\subsection{Datasets}
We train and evaluate our proposed approach on H2O~\cite{kwonH2OTwoHands2021} and FPHA~\cite{Garcia-Hernando_2018_CVPR}.

\noindent{\textbf{The H2O Dataset.}} \qquad H2O dataset \cite{kwonH2OTwoHands2021} is acquired in indoor settings in which the subjects interact with objects using both of their hands. It includes more than 570K RGB-D frames and 933 trimmed action clips. The dataset features 36 distinct action classes, which are combinations of 8 object classes and 11 verb classes. The dataset also includes annotated accurate ground-truth data for left and right hand pose, 6D object pose, camera pose, object labels, action labels, and object meshes, with which we can derive ground-truth contact and distant points for the training of the contact-ware module. 

\medskip
\noindent{\textbf{The FPHA Dataset.}} \qquad FPHA dataset \cite{Garcia-Hernando_2018_CVPR} studies the use of 3D hand poses of only one hand to recognize first-person dynamic hand actions interacting with 3D objects. It includes RGB-D video sequences comprised of more than 100K frames of 45 daily hand action categories, involving 26 different objects in several hand configurations. The dataset also includes annotated accurate ground-truth data for right hand pose, 6D object pose, viewpoints, object labels, action labels, and object meshes. However, 6D object pose and object meshes are only available in 4 objects related to 10 action classes in this dataset, and therefore, we use the subset where these data exist for training and testing, following \cite{Tekin_2019_CVPR, Cho_2023_CVPR}. 

Note that due to the size of the dataset and the availability of two hands, we decide to set up our experiment mainly on the H2O dataset.
By following this, for the simplicity of illustration, the dimensions of variables above and in the below subsections are only for the H2O dataset, adjustments should be made to adapt to FPHA or other datasets.

\subsection{Implementation Details}
As we show in our ablation study, we employ MLP-based networks for both networks $f$ and $g$ described in Section~\ref{sec:CaSAR}. Both of them have 2 hidden layers, with ReLU being the activation function for hidden layers and Sigmoid being the activation function for the last layer. The number of nodes in each hidden layer is 256, 5000 for $f$ and $g$, respectively. Note that the input, $x_i$, is flattened before feeding into the networks.

In practice, we first train the contact-aware module $f$ by setting $\lambda =0$ and then freeze its weight to predict the action recognition module $g$. The reason is that the train set $\mathcal{T}^C$ does not require action labels, therefore, the network $f$ can be trained on more samples. In addition, it removes the need of tuning the hyperparameter $\lambda$ in Eq.~\ref{eq:loss_total}. The model is trained with the Adam optimizer \cite{kingmaAdamMethodStochastic2017} with the learning rate schedule that follows a step-wise decay strategy. Specifically, we start with $1 \times 10^{-4}$, and every 20 epoch we reduce it by 30\% and train for a total number of 100 epochs. We train the network $g$ for 600 epochs, with the initial learning rate starting at $1 \times 10^{-5}$ and we reduce it by 30\% for every 200 epochs.

\medskip
\noindent{\textbf{Uniform frame length.}} \qquad The length of the frames for each action exhibits considerable variability. To match the input size of the networks, it is necessary to standardize the frame length across all actions. Empirically, we set a uniform length of $N_f = 32$ frames as we observed this number as the optimal number of frames to extract the requisite information. When we encounter video segments exceeding the length, we select  $N_f$ frames uniformly from the original video. For video segments shorter than this length, we use a repeat strategy to duplicate frames
to pad it up to the total length, which works best in our case. We also tried different strategies, such as zero padding, but did not bring improvement compared to the current strategy.

\begin{figure}
\begin{center}
    \begin{subfigure}[b]{0.4\columnwidth}
        \includegraphics[width=\linewidth]{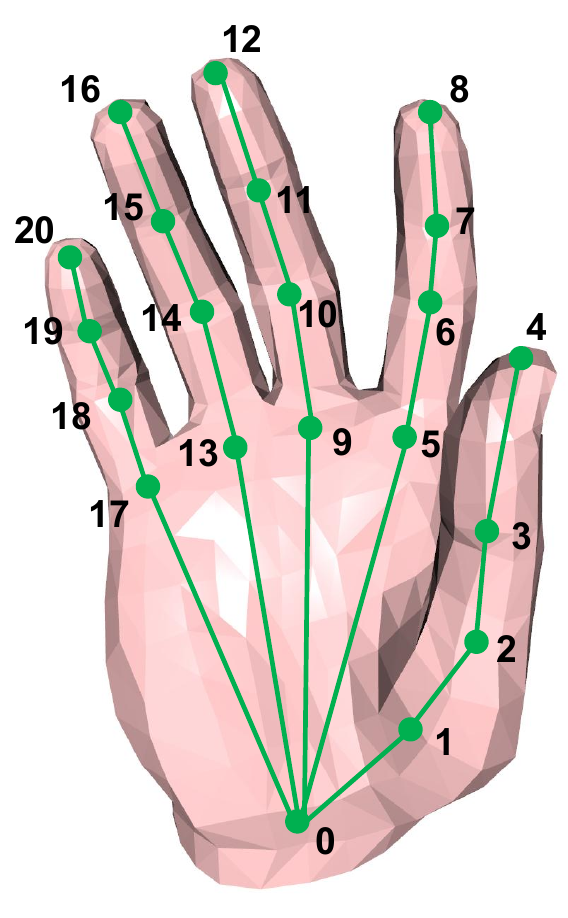}
        \caption{}
        \label{fig: hand}
        \end{subfigure}
    \hspace{10mm}
    \begin{subfigure}[b]{0.4\columnwidth}
        \includegraphics[width=\linewidth]{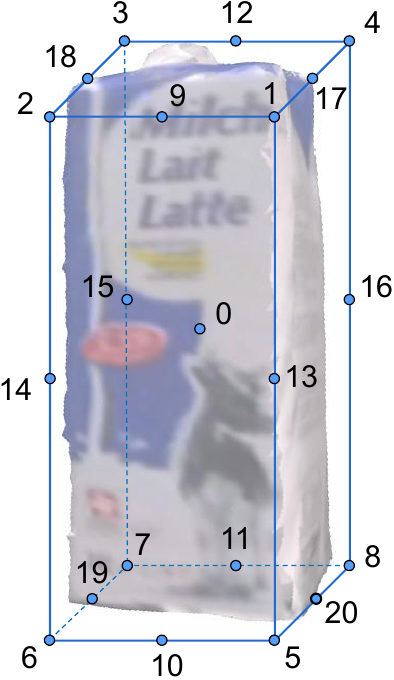}
        \caption{}
        \label{fig:object}
    \end{subfigure}
\end{center}
  
  \caption{Model and enumerated points for (a) hand and joints, (b) 8-corner bounding box.}
  \label{hand&object} 
\end{figure}

\medskip
\noindent{\textbf{Hand pose annotation.}} \qquad H2O \cite{kwonH2OTwoHands2021} dataset provides MANO \cite{MANO:SIGGRAPHASIA:2017} hand fits for both left and right hand, as shown in Figure \ref{fig: hand}. The order of the joints is fixed. In each frame, the 3D coordinates of the 42 joints of two hands are also given; we conclude them in the vector $h$ as follows:
\begin{equation}
    \mathbf{h} = [\mathbf{h}^{\text{Left}}, \mathbf{h}^{\text{Right}}] \subset \mathbb{R}^{126}
\end{equation}
Where $\mathbf{h}^{\text{Left}}, \mathbf{h}^{\text{Right}} \subset \mathbb{R}^{63}$ denote the 3D coordinates of 21 joints on the left and right hands.

\medskip
\noindent{\textbf{Object pose annotation.}} \qquad H2O \cite{kwonH2OTwoHands2021} dataset provides high-quality object meshes. With the provided $4 \times 4$ homogeneous transformation matrix in each frame, we can obtain the position of the object at the corresponding moment. Additionally, the label of the object, and the 3D coordinates of 21 points (1 centre, 8 corners, 12 mid-edge points, as illustrated in Figure \ref{fig:object}) of the 8-corner bounding box in each frame are also included in the dataset. The order of the points is fixed. We conclude them in the object label vector $\mathbf{l} \subset \mathbb{R}^{8}$ and the object pose vector $\mathbf{o} \subset \mathbb{R}^{63}$.

\subsection{Comparison Framework}

\noindent{\textbf{Comparison to State-of-the-Art Baselines}} \qquad We compare our skeletal action recognition result to the state-of-the-art methods in Table~\ref{tab:sota}. With 8-corner bounding boxes as object pose representation, 3D coordinates of joints as hand pose representation, and contact points and distant points as additional input features, our CaSAR model achieves an accuracy of $91.3\%$ and $98.4\%$ on top-1 action class that matches the object class, which are the best accuracy among SOTA methods on the H2O, and FPHA datasets, respectively.
It is notable that, in contrast to H2OTR~\cite{Cho_2023_CVPR} that relies on object meshes, ours does not require mesh at inference time. 

\begin{table}[h]
    \centering
    \begin{tabular}{@{}llcc@{}}
        \toprule
        \multirow{2}{*}{Model} & \multirow{2}{*}{Modality} & H2O & FPHA\\
        \cmidrule{3-4}
        ~ & & Acc.(\%)& Acc.(\%)\\
        \Xhline{0.5pt}
        C2D\cite{wang2018nonlocal}& RGB & 70.7 & - \\
        I3D\cite{carreira2018quo}& RGB & 75.2 & - \\
        SlowFast\cite{feichtenhoferSlowFastNetworksVideo2019a}& RGB & 77.7 & - \\
        Tekin et al.\cite{Tekin_2019_CVPR} & RGB & 68.9 & 97.0 \\
        TA-GCN~\cite{kwonH2OTwoHands2021}& *Skeleton & 79.3 & - \\
        Wen et al. \cite{wen2023hierarchical}& RGB & 86.4 & - \\
        H2OTR \cite{Cho_2023_CVPR}& *Skeleton+contact & 90.9 & \textbf{98.4}  \\
        \textbf{Ours} & Skeleton+contact & \textbf{91.3} & \textbf{98.4} \\
        \bottomrule
    \end{tabular}
    \caption{Comparison to state-of-the-art methods for action recognition on test sets of H2O and FPHA. Note that our method uses ground-truth skeleton information directly from the H2O and FPHA datasets. * indicates predicted skeletons.}
    \label{tab:sota}
\end{table}

\begin{figure}
    \centering
    \includegraphics[width=0.5\textwidth]{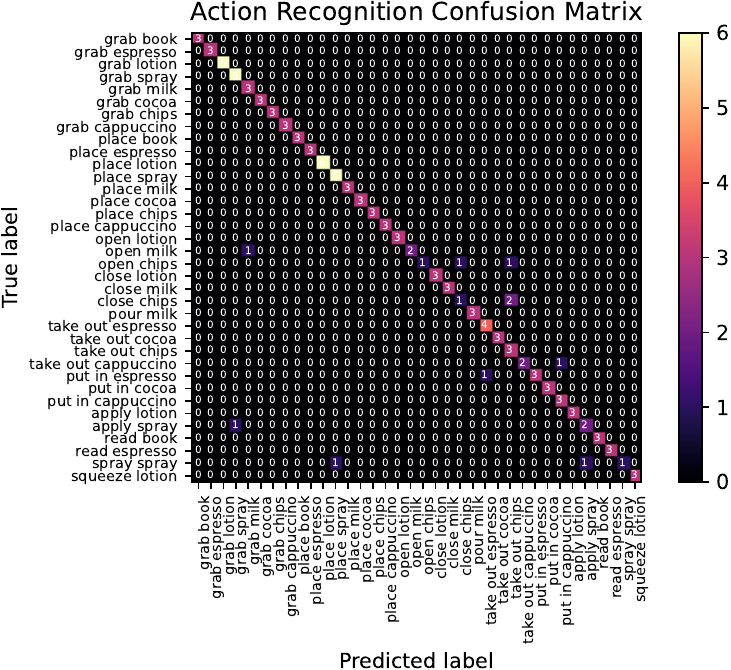}
    \caption{Confusion matrix for action recognition of H2O.}
    \label{confusion_matrix}
\end{figure}

\begin{figure*}
    \centering
    \begin{subfigure}[b]{0.65\textwidth}
        \includegraphics[width=\textwidth]{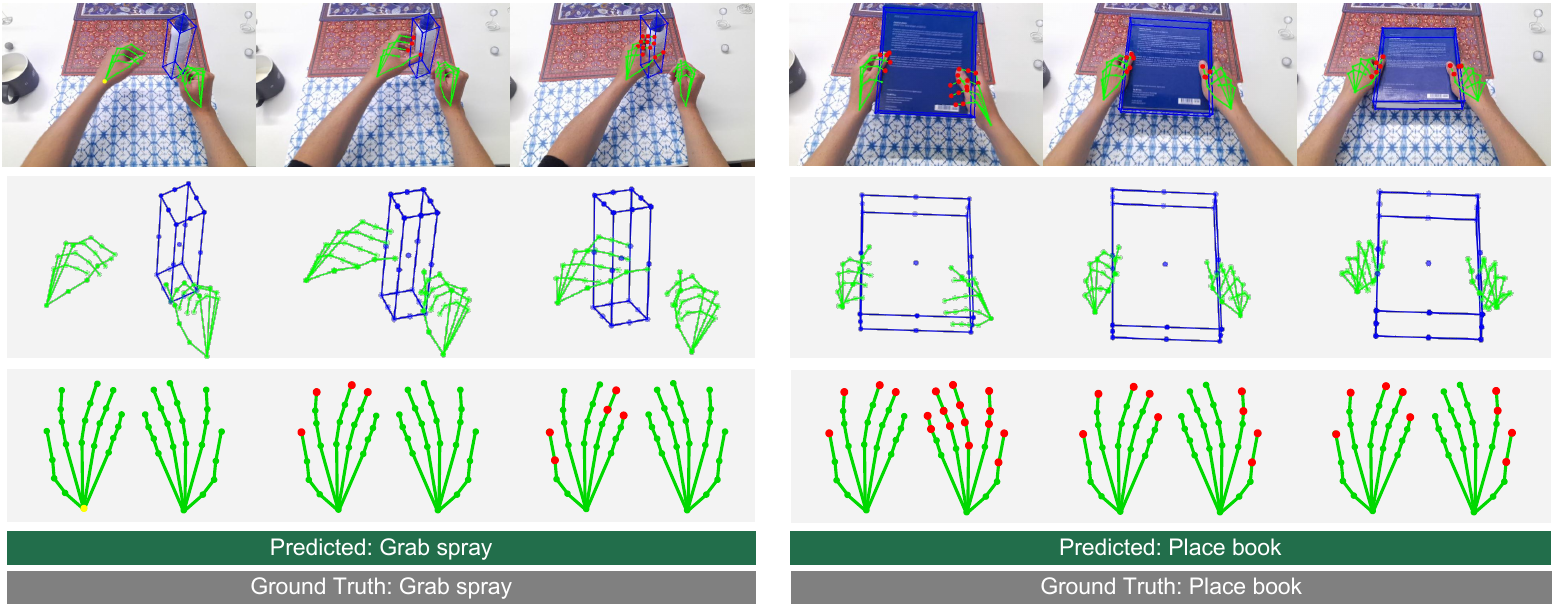}
        \caption{}
        \label{fig: ar_h2o_sucess}
        \end{subfigure}
    \hspace{3mm}
    \begin{subfigure}[b]{0.32\textwidth}
        \includegraphics[width=\textwidth]{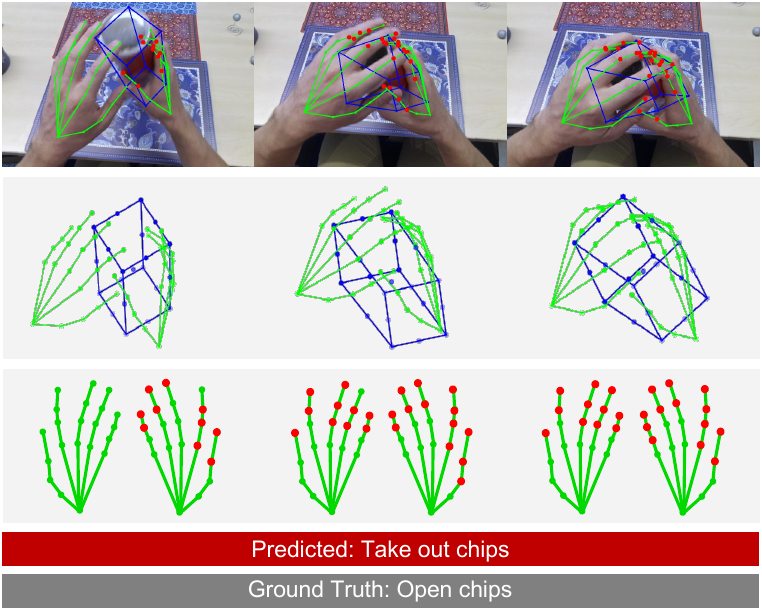}
        \caption{}
        \label{fig:ar_h2o_fail}
    \end{subfigure}
    \caption{Examples of success cases (a) and a failure case (b) of action recognition on the H2O dataset predicted by our CaSAR. (Row 1) Image frame, (Row 2) Input skeleton frame, (Row 3) Contact map and distant map, (Row 4) Estimated action classes, (Row 5) Ground truth action classes.}

    \label{fig:ac_h2o}
\end{figure*}

\begin{figure*}
    \centering
    \begin{subfigure}[b]{0.65\textwidth}
        \includegraphics[width=\textwidth]{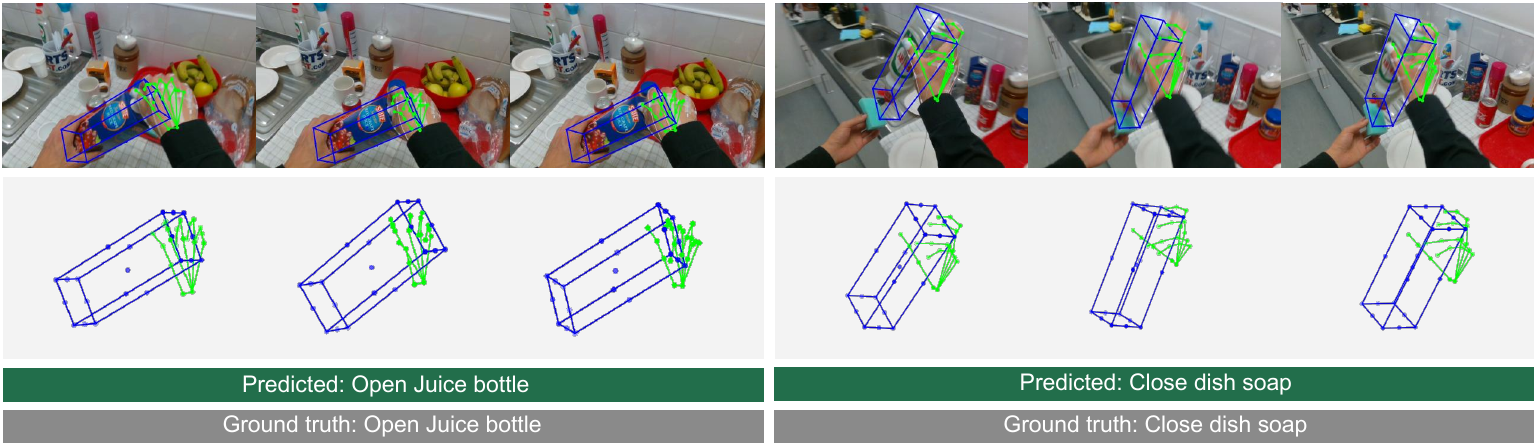}
        \caption{}
        \label{fig: ar_fpha_sucess}
        \end{subfigure}
    \hspace{3mm}
    \begin{subfigure}[b]{0.32\textwidth}
        \includegraphics[width=\textwidth]{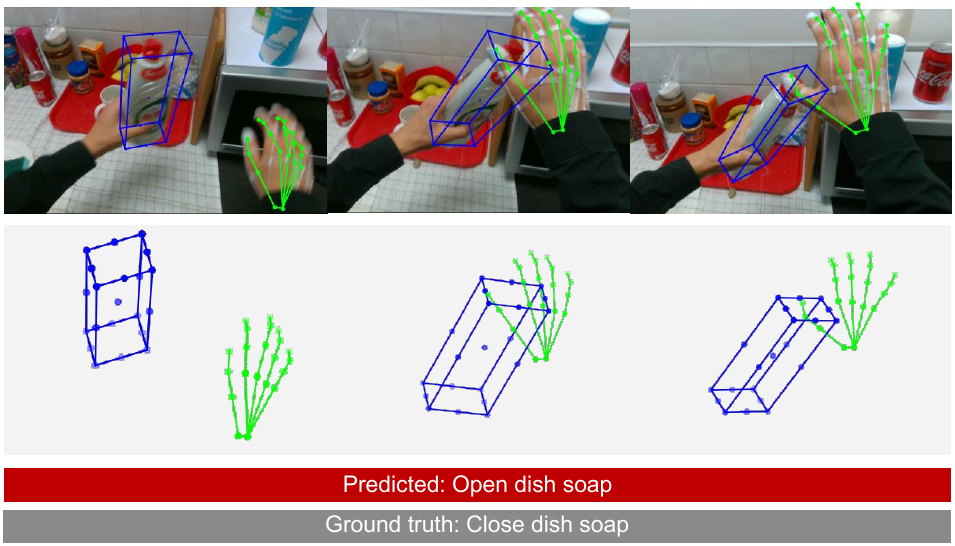}
        \caption{}
        \label{fig:ar_hfpha_fail}
    \end{subfigure}
    \caption{Examples of success cases (a) and a failure case (b) of action recognition on the FPHA dataset predicted by our method, CaSAR. (Row~1) Image frame, (Row~2) Input skeleton frame, (Row~3) Estimated action classes, (Row~4) Ground truth action classes.}
    \label{fig:ac_fpha}
\end{figure*}

\noindent{\textbf{Qualitative and Quantitative Results}} \qquad  In Figure \ref{confusion_matrix}, we show the confusion matrix for the ground truth and predicted action labels. While our model achieves high accuracy in distinguishing among different classes, the prediction accuracy of \texttt{Chips} related actions is lower than other classes. This can be explained by the lower accuracy of contact point prediction of the object \texttt{Chips} as shown in Table~\ref{tab:first-stage}. Also, the misclassification of the \emph {``spray Spray"} action is related to the low distant point prediction accuracy of the object \texttt{Spray}.  Figure \ref{fig:ac_h2o} and \ref{fig:ac_fpha} show qualitative results on both success and failure cases of our proposed approach on the H2O and FPHA datasets, respectively.

\begin{figure*}
\begin{center}
    \begin{subfigure}[b]{0.63\textwidth}
        \includegraphics[width=\textwidth]{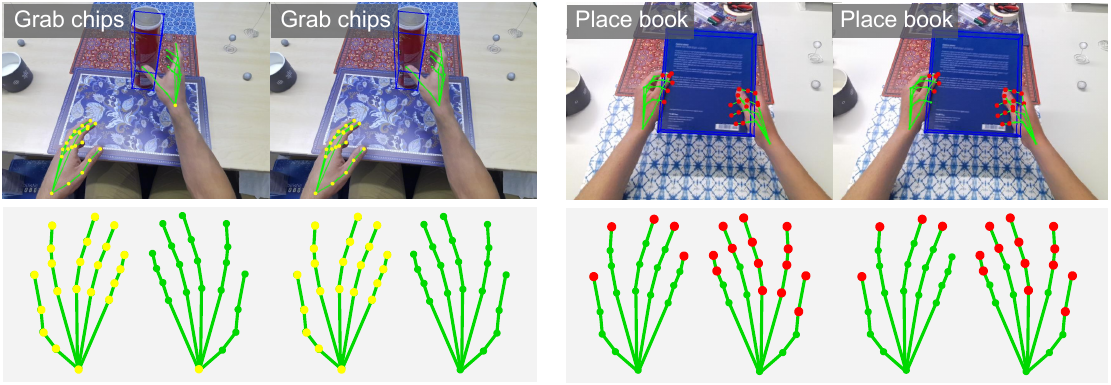}
        \caption{}
        \label{fig: contact_s_h2o}
        \end{subfigure}
    \hspace{5mm}
    \begin{subfigure}[b]{0.322\textwidth}
        \includegraphics[width=\textwidth]{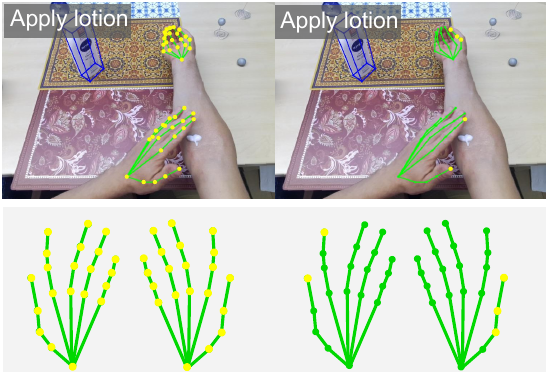}
        \caption{}
        \label{fig:contact_f_h20}
    \end{subfigure}
    \caption{Examples of success cases (a) and a failure case (b) of contact-map and distant map prediction (right), with ground truth (left) on H2O dataset predicted by the contact-aware module of our CaSAR.  }
    \label{fig:contact_h2o}
\end{center}
\end{figure*}

\noindent{\textbf{Contact-map evaluation.}} \qquad 
Table~\ref{tab:first-stage} shows the prediction accuracy of contact-aware module $f$, including both contact and distant points, for different objects. We can observe that predicting distant points is more accurate on average. This is expected as contact points have much harder constraints to be predicted compared to distant points.
Upon analyzing individual objects closely, we observe that the accuracy of predicting contact points varies across different object types. Particularly, the highest accuracy in contact point prediction is attained for objects labeled as \texttt{Book}, \texttt{Espresso}, \texttt{Cocoa}, and \texttt{Cappuccino}, while this precision diminishes for the \texttt{Chips} object, which possesses a cylindrical shape. This discrepancy in accuracy could be attributed to the distinct shapes of these objects. Objects that exhibit cube-like geometries align closely with the concept of an 8-corner bounding box, thereby facilitating accurate contact point estimation. Conversely, the \texttt{Chips} object's cylindrical form complicates the process of deriving contact maps from an 8-corner bounding box representation. Furthermore, the \texttt{Chips} object involves actions like \emph{``take out Chips"}, wherein a substantial part of the hand penetrates the object's interior. This contrasts with interactions concerning other objects, primarily occurring on the object's surface. Consequently, predicting contact points for the \texttt{Chips} object becomes more challenging. Figure~\ref{fig:contact_h2o} shows some examples of the predicted contact-map.

\begin{table}[h]
    \centering
    \begin{tabular}{@{}lcc@{}}
        \toprule
        Object &  Contact acc. (\%) & Distant acc. (\%) \\
        \midrule
        Book & 92.0 & 99.8 \\
        Espresso & 94.1 & 98.9 \\
        Lotion & 90.1 & 91.4 \\
        Spray & 90.8 & 90.9 \\
        Milk & 90.0 & 98.1 \\
        Cocoa & 95.1 & 98.3 \\
        Chips & 83.5 & 99.4 \\
        Cappuccino & 91.9 & 99.8 \\
        \midrule
        Average & 90.8 & 96.3 \\
        \bottomrule

    \end{tabular}
    \caption{Contact-map prediction. We categorize each contact point and distant point prediction accuracy by objects. \texttt{Book} achieves the highest in both contact and distance accuracy, whereas \texttt{Chips} produces the lowest contact accuracy and \texttt{Spray} produces the lowest distance accuracy.}
    \label{tab:first-stage}
\end{table}

\begin{table}[h]\scriptsize
    \centering
    \begin{tabular}{@{}lccccl@{}}
        \toprule
        Model & Obj repre. & Contact p. & Distant p. & Acc. (\%)\\
        \midrule
        \multirow{1}{*}{LSTM baseline} & bbox & $\cdot$  & $\cdot$ &  72.3\\
        \midrule
        \multirow{2}{*}{MLP baseline} %
        ~ & Poisson & $\cdot$ & $\cdot$ &  79.8\\
          ~ & bbox  & $\cdot$ & $\cdot$ &  84.7\\
        \midrule
        \multirow{3}{*}{CaSAR} %
        ~ & bbox  &  \ding{51} & $\cdot$ &  90.1\\
         ~ & bbox  & $\cdot$  & \ding{51} &  84.7\\
         ~ & bbox   &  \ding{51} & \ding{51} &  \textbf{91.3}\\
        \bottomrule

    \end{tabular}
    \caption{Ablation study. We run two baselines, LSTM and MLP, as well as different object representations Poisson sampling and bounding boxes to understand better representation for our task. Then, we ablate contact-map to know the effectiveness of constant points and distant points for action recognition. Note that we set up the ablation study on the H2O dataset.
    }
    \label{tab:CaSAR}
\end{table}

\subsection{Ablation Study}
We discuss here the results of experiments we perform using H2O to describe better our method.

\noindent{\textbf{Baselines.}} \qquad
We introduce two baseline network architectures in our study. Firstly, we incorporate LSTM \cite{Hochreiter} as a baseline model due to its established efficacy in handling sequential data, demonstrated in domains like action recognition~\cite{Garcia-Hernando_2018_CVPR}. Secondly, we employ MLP \cite{haykin1994neural} as the other baseline. The comparison between the vanilla MLP and LSTM models is shown in Table~\ref{tab:CaSAR}. Particularly, the MLP model demonstrates superior performance to LSTM. This can be attributed to the proficiency of MLPs, as they are generally good at capturing simpler patterns and relationships within data, when the dataset is small. The limited size of the data also makes the task not suitable for more complex network architectures such as Transformers~\cite{vaswani2017attention}.  In addition, LSTM, with its ability to capture long-range dependencies, can sometimes lead to overfitting when applied to tasks where shorter dependencies are sufficient. To this end, we choose our backbone model as MLP because of its interpretability and the fact that the MLP baseline produces a better result than the LSTM baseline.

\noindent{\textbf{The impact of Contact and Distant Points.}}  \qquad Last rows of Table~\ref{tab:CaSAR} show the impact  of contact-map. Using contact points improves the accuracy significantly by 5.4\%, and using distant points brings an additional improvement of 1.2\%. 
Notably, our contact-map module plays a substantial role, accounting for 6.6\% of the overall performance improvement. This outcome underscores the critical role that spatial information, in this case, contact-map, plays in enhancing action recognition capabilities.

\noindent{\textbf{Object Representation.}}  \qquad
We also evaluate the impact of object representation on the model's performance. The 8-corner bounding box offers insights into object translation and orientation, yet falls short in providing intricate surface and shape. Object mesh representation, for example, can impart superior geometric insights, which could be potentially essential for understanding hand-object interaction in actions. However, this requires the model to learn an implicit understanding of 3D interaction, which makes the task difficult as it requires a complex model and a large amount of data. We, therefore, conduct a comparative analysis of object representation techniques by providing a subset of object mesh using Poisson sampling. Table~\ref{tab:CaSAR} confirms that bounding box representation brings a notable 4.9\% performance boost in the baseline MLP model.

\subsection{Computation Times}
We implement our network using PyTorch \cite{NEURIPS2019_bdbca288} and the code runs on an Intel i7-12700H CPU with a NVIDIA GeForce RTX 3050 Ti GPU. It takes 25 ms for the contact-map and action recognition at inference time for each action.

\section{Conclusions}

In this paper, we propose a contact-aware action recognition framework, CaSAR, that can predict hand-object contact relations which does not require 3D meshes of the object during the testing time. The framework can learn contact relations explicitly from hand joints and object bounding boxes and geometric shapes implicitly. We further propose a novel concept, contact-map consisting of contact points and distant points, that can significantly improve the performance of action recognition.

To calculate contact maps, our method relies on hand poses and object poses from the datasets. Wrong calculation of contact-map will result in inaccuracies in contact-map prediction action recognition. Also, current datasets with ground truth 3D hand pose and 6D object pose are limited in size compared to large-scale egocentric video datasets such as Ego4D~\cite{grauman2022ego4d}, HoloAssist~\cite{HoloAssit}, and EpicKitchen~\cite{Damen2021PAMI}. It would be interesting to collect a large-scale dataset to obtain more generalized contact-map. End-to-end action recognition by estimating hand poses, object poses and contact-map from RGB images directly would also be an interesting future direction.

{\small
\bibliographystyle{ieee_fullname}
\bibliography{egbib}
}

\end{document}